\documentclass[preprint,12pt]{elsarticle}

\usepackage{amsmath,amssymb,amsfonts}
\usepackage{algorithmic}
\usepackage{graphicx,color}
\usepackage{textcomp}
\usepackage{xcolor}
\usepackage{hyperref}
\hypersetup{hidelinks=true}
\usepackage{algorithm,algorithmic}
\usepackage[font=footnotesize]{caption}

\renewcommand{\emph}[1]{\textit{#1}}

\newcommand{\change}[1]{\ensuremath{\operatorname{#1}}}
\newcommand{\MAT}{ [ \begin{array}}  
\newcommand{\mat}{\end{array}  ]}
\newtheorem{Definition}{Definition}[section]

\newtheorem{Q}{Question}[section]

\def \minimize {\operatorname*{minimize}}

\def \st {\operatorname*{s.t. }}
\usepackage{subcaption}
\usepackage{amsmath,amssymb} 
\usepackage{bm}
\usepackage{float}

\def\x{{\mathbf x}}

\def\r{ \mathbf{r}}

\def\g{{\mathbf g}}

\def\B{{\mathbf B}}

\def\T{{\mathrm T}}

\def\RR{{\mathbb R}}
\def\NN{{\mathbb N}}

\def\bb{{\mathbf b}}

\newcommand{\paren}[1]{\left(#1\right)}

\newcommand{\vct}[1]{\mathbf{#1}}
\newcommand{\tTx}[1]{\mathbf{#1}}

\newcommand{\bmtx}[1]{\mathbf{#1}}
\newcommand{\inner}[1]{\left<#1\right>}

\def \vec       {\operatorname*{vec}}

\newcommand{\vb}{\vct{b}}

\newcommand{\vm}{\vct{m}}

\newcommand{\vx}{\vct{x}}
\newcommand{\vy}{\vct{y}}

\newcommand{\mB}{\tTx{B}}

\newcommand{\mM}{\tTx{M}}

\newcommand{\mQ}{\tTx{Q}}

\newcommand{\mX}{\tTx{X}}

\newcommand{\mId}{\bmtx{I}}

\newcommand{\mzero}{{\bf 0}}

\def \st {\operatorname*{s.t.\ }}

\newcommand{\tensor}[1]{\boldsymbol{\mathcal{#1}}}

\def \tB {\tensor{B}}

\def \tG {\tensor{G}}

\def \tX {\tensor{X}}

\journal{Signal Processing}

\begin{document}

\begin{frontmatter}



\title{A Muon-Accelerated Algorithm for Low Separation Rank Tensor Generalized Linear Models}


\author[1]{Xiao Liang}
\ead{liangx@iastate.edu}

\author[1]{Shuang Li}
\ead{lishuang@iastate.edu}

\affiliation[1]{organization={Department of Electrical and Computer Engineering, Iowa State University},
                city={Ames},
                postcode={50014}, 
                state={Iowa},
                country={USA}}

\cortext[cor1]{Corresponding author}                

\begin{abstract}
Tensor-valued data arise naturally in multidimensional signal and imaging problems, such as biomedical imaging. When incorporated into generalized linear models (GLMs), naive vectorization can destroy their multi-way structure and lead to high-dimensional, ill-posed estimation. To address this challenge, Low Separation Rank (LSR) decompositions reduce model complexity by imposing low-rank multilinear structure on the coefficient tensor. A representative approach for estimating LSR-based tensor GLMs (LSR-TGLMs) is the Low Separation Rank Tensor Regression (LSRTR) algorithm, which adopts block coordinate descent and enforces orthogonality of the factor matrices through repeated QR-based projections. However, the repeated projection steps can be computationally demanding and slow convergence. Motivated by the need for scalable estimation and classification from such data, we propose LSRTR-M, which incorporates Muon (MomentUm Orthogonalized by Newton-Schulz) updates into the LSRTR framework. Specifically, LSRTR-M preserves the original block coordinate scheme while replacing the projection-based factor updates with Muon steps. Across synthetic linear, logistic, and Poisson LSR-TGLMs, LSRTR-M converges faster in both iteration count and wall-clock time, while achieving lower normalized estimation and prediction errors. On the Vessel MNIST 3D task, it further improves computational efficiency while maintaining competitive classification performance.
\end{abstract}


\end{frontmatter}



\section{Introduction}

Generalized Linear Models (GLMs) form a broad class of regression models, including linear regression, logistic regression, and Poisson regression~\cite{mccullagh2019generalized, hastie2017generalized, nelder1972generalized}. They generalize ordinary linear regression by allowing the response variable to follow a distribution from the exponential family. The conditional distribution of the response is often characterized by a linear predictor and a strictly monotonic, invertible link function that relates the covariates to the conditional mean of the response variable. The model parameters are typically estimated using maximum likelihood, that is, by minimizing the empirical negative log-likelihood over a set of training samples.

GLMs are widely used in signal and imaging applications for estimation, detection, and classification~\cite{tan2012logistic, li2018tucker, zhang2016decomposition}. This setting becomes especially important when observations are naturally represented as tensors, as in biomedical imaging~\cite{zhou2013tensor,taki2023structured}. In such cases, preserving multi-way structure can improve both statistical efficiency and computational tractability relative to vectorization, making low-rank tensor models a natural tool for regression and classification from tensor-valued measurements.

Two of the most widely used tensor decompositions are the CANDECOMP/PARAFAC (CP) decomposition and Tucker decomposition~\cite{kolda2009tensor, sidiropoulos2017tensor, tokcan2025tensor}. These structures have been incorporated into tensor regression and tensor GLMs, leading to efficient algorithms for CP- and Tucker-structured coefficient tensors recovery~\cite{li2018tucker, zhou2013tensor, ahmed2020tensor}. By restricting the parameter space to structured low-rank tensors, such approaches substantially reduce the effective model complexity compared to fully unstructured formulations. More recently, Taki et al. introduced a Low Separation Rank (LSR) decomposition for the coefficient tensor in TGLMs~\cite{taki2023structured, taki2024low}. This LSR model generalizes the Tucker decomposition and can also be viewed as a special case of the Block Tensor Decomposition (BTD)~\cite{de2008decompositions_two, rontogiannis2021block}.
Under the LSR parameterization, the coefficient tensor is represented as a sum of structured Tucker-type components, each described by mode-wise factor matrices and a shared core tensor. To ensure identifiability and numerical stability, these factor matrices are typically constrained to have orthonormal columns.

To leverage the statistical advantages of the LSR structure in TGLMs, Taki et al.~\cite{taki2023structured} proposed the Low Separation Rank Tensor Regression (LSRTR) algorithm, a block coordinate descent (BCD) method for estimating the coefficient tensor. This algorithm alternately updates the factor matrices and the shared core tensor while enforcing orthogonality constraints on the factor matrices. Although this BCD framework is computationally appealing and well aligned with the structure of the LSR decomposition, each factor update requires repeated gradient steps followed by QR-based orthogonal projections to enforce orthogonality constraints on the factor matrices. As a result, the per-iteration cost may be significantly influenced by the repeated orthogonalization operations, especially in high-dimensional settings. For large tensor-valued datasets, however, the repeated QR projections in LSRTR can become a practical computational bottleneck. Our aim is therefore to develop a more scalable solver for LSR-TGLMs that remains suitable for multidimensional signal and imaging data analysis.

The recently introduced Muon (MomentUm Orthogonalized by Newton-Schulz) optimizer~\cite{jordan2024muon} has attracted attention for its computational efficiency in large-scale optimization settings~\cite{pethick2025training, liu2025muon}. The central idea behind Muon is to maintain a momentum accumulator and orthogonalize the matrix update direction via an approximate inverse square root (equivalently, a polar-type map), computed efficiently using Newton-Schulz iterations. Recent studies have begun to establish convergence analyses and adaptive variants for such orthogonalized updates~\cite{li2025note, zhang2025adagrad}. Empirically, this design has been observed to improve optimization stability and computational efficiency compared to conventional projection-based gradient schemes.

In this work, we integrate orthogonalized-momentum updates into the LSR-TGLM framework by introducing the LSRTR-M algorithm, a Muon-accelerated variant of LSRTR. The proposed method preserves the original block coordinate descent structure and the core-tensor update, while modifying the factor-matrix subproblems. Specifically, instead of performing explicit QR-based projections after each gradient step, we update each factor block using a momentum-based orthogonalized direction, computed via a lightweight Newton-Schulz procedure. This replacement avoids explicit projection steps and provides a computationally efficient update rule for the factor matrices.

The main contributions of this work are summarized as follows.
\begin{itemize}
\item We propose the LSRTR-M algorithm, a Muon-accelerated variant of LSRTR for LSR-TGLMs.
\item We show on synthetic TGLM problems, including linear, logistic, and Poisson models, that LSRTR-M yields faster estimation of LSR-TGLMs, achieving lower normalized estimation and prediction errors with fewer iterations and reduced wall-clock time.
\item We validate the practical relevance of the proposed method on the Vessel MNIST 3D classification task, demonstrating improved computational efficiency and competitive or better predictive performance on tensor-valued biomedical imaging data~\cite{yang2021medmnist,yang2020intra, yang2023medmnist}.
\end{itemize}

The remainder of this paper is organized as follows. We introduce some key notations and definitions in Section~\ref{sec:Preliminaries}. The TGLMs and LSR-TGLM estimation problem are mathematically formulated in Section~\ref{sec:Problem Statement}. We present the LSRTR-M algorithm, detailing the BCD framework and Muon-style orthogonalized-momentum factor updates in Section~\ref{sec:Algorithm design} and report experimental results on synthetic TGLMs and the Vessel MNIST 3D benchmark in Section~\ref{sec:Experiment}. Finally, we conclude this work and discuss future directions in Section~\ref{sec:Conclusion}.  

\section{Preliminaries}
\label{sec:Preliminaries}

\textbf{Notation.} 
Throughout this paper, we use bold lowercase letters (e.g., $\vx$), bold uppercase letters (e.g., $\mX$), and bold calligraphic letters (e.g., $\tX$) to represent vectors, matrices, and tensors, respectively. For a tensor $\tX$, $\vx \triangleq \vec(\tX)$ denotes its column-wise vectorization.
For a positive integer $K$, we define $[K] \triangleq \{ 1, 2, \dots, K \}$.
For two matrices $\mX_1$ and $\mX_2$, $\mX_1 \otimes \mX_2$ denotes their Kronecker product.   
The inner product between two vectors, matrices, or tensors is denoted by $\langle \cdot, \cdot \rangle$. 
The mode-$k$ matricization of a tensor $\tX$ is denoted by $\tX_{(k)}$.  
Given a matrix $\mB$, the mode-$k$ multiplication of $\tX$ by $\mB$ is denoted by $\tX\times_k \mB$. 

We start from the classical GLMs with vector-valued covariates before extending the framework to tensor-structured models (TGLMs).

\begin{Definition}[Vector-structured GLMs~\cite{taki2023structured}]
\label{def:vector_glm}
Consider an observation $y\in \RR$, a covariate vector $\vx\in\RR^{m}$, and unknown parameters $\bb\in\RR^{m}$ (regression coefficient vector) and $z\in \RR$ (bias).
Let $y$ be a response variable generated from a distribution in the exponential family with probability mass/density function as follows:
\begin{equation}
\label{eq:exp_family_def1}
  \mathbb{P}(y ; \eta) = b(y)\exp\!\big(\eta\,\T(y)-a(\eta)\big),  
\end{equation}
where $\T(y)$ is the sufficient statistic. $a(\eta)$ is the log-partition function with $\eta$ being the natural parameter, and $b(y)$ is the base measure (carrier measure) of the exponential family distribution.
The GLM links the conditional mean $\mu$ to the natural parameter $\eta$ via a strictly monotonic link function:
\[
g(\mu)=\eta \triangleq \inner{\bb,\vx}+z,
\]
where the natural parameter $\eta$ serves as the linear predictor. The conditional mean $\mu \;=\; \mathbb{E}[y \mid \vx] \;=\; g^{-1}(\eta)$, $g(\cdot)$ is a strictly monotonic, invertible link function, and $z$ is the bias.
Note that the scalar response $y$ is related to the linear predictor via $g(\cdot)$, and its distribution is only affected by the linear combination $\inner{\bb,\vx}+z$.
\end{Definition}


Linear, logistic, and Poisson regression arise as special cases of GLMs.
In linear regression, assuming a Gaussian response with the identity link, one may take
$\T(y)=y$, $a(\eta)=\eta^2/2$, and
$b(y)=(2\pi)^{-1/2}\exp(-y^2/2)$, which implies that the conditional mean satisfies
$\mu=\inner{\bb,\x}+z$.
In logistic regression, where $y\in\{0,1\}$ follows a Bernoulli distribution and the logit link
$g(\mu)=\log\!\paren{\frac{\mu}{1-\mu}}$ is used, we have
$\T(y)=y$, $a(\eta)=\log(1+e^\eta)$, and $b(y)=1$, from which it follows that the conditional mean is
$\mu=\sigma(\eta)=\frac{1}{1+\exp(-\eta)}$.
In Poisson regression, where $y\in\mathbb{N}$ follows a Poisson distribution and the log link
$g(\mu)=\log(\mu)$ is adopted, we have
$\T(y)=y$, $a(\eta)=e^\eta$, and $b(y)=1/y!$, which yields the conditional mean 
$\mu=\exp(\eta)$.

When the covariates are tensors, the regression coefficient naturally becomes a tensor parameter. Vectorizing both the response model and the coefficient tensor leads to a high-dimensional linear predictor involving Kronecker-structured operators. To reduce model complexity and exploit multi-mode structure, it is natural to impose structural constraints on the coefficient tensor. One such structural assumption is Low Separation Rank (LSR), which generalizes single Kronecker-structured parameterizations~\cite{taki2023structured,tsiligkaridis2013covariance, van1993approximation}.

\begin{Definition}[Matrix Separation Rank~\cite{tsiligkaridis2013covariance}]
Let $K\in\NN$. Let $\vm=(m_1,\ldots,m_K)\in\NN^K$ and
$\r=(r_1,\ldots,r_K)\in\NN^K$.
Denote $\tilde m = \prod_{k\in[K]} m_k$ and $\tilde r = \prod_{k\in[K]} r_k$.
For a matrix $\mB\in\RR^{\tilde m\times \tilde r}$, its separation rank, denoted by $\change{sep\_rank}(\mB)$, is the minimum number $S$ of $K$-order Kronecker-structured matrices such that
\begin{equation}
\mathbf{B}=\sum_{s \in[S]} \mathbf{B}_{(1, s)} \otimes \cdots \otimes \mathbf{B}_{(K, s)},
\label{eq:sep-rank}
\end{equation}
where $\mB_{(k,s)}\in\RR^{m_k\times r_k}$ for all $k\in[K]$ and $s\in[S]$.
If $\change{sep\_rank}(\mB)$ is small, $\mB$ is said to have low separation rank.
\end{Definition}

We now extend the notion of separation rank from matrices to tensors and formally define the LSR tensor decomposition.

\begin{Definition}[LSR Tensor Decomposition] \label{def:LSR_tensor}
Let $\tB$ be a $K$-mode tensor (of size $m_1\times\cdots\times m_K$).
The rank-$(r_1,\ldots,r_K)$ LSR decomposition with separation rank $S$ decomposes $\tB$ as 
\begin{equation}
\label{LSR_defn}
\tB \;=\; \sum_{s\in[S]} \tG \times_1 \B_{(1,s)} \times_2 \cdots \times_K \B_{(K,s)}, 
\end{equation}
where the core tensor $\tG \in \RR^{r_1\times \cdots \times r_K}$ is shared across all components, and
$\B_{(k,s)} \in \RR^{m_k\times r_k},\ k\in[K],\ s\in[S]$ denote the Kronecker-structured factor matrices.
Equivalently, defining $\g \triangleq \vec(\tG)$ and $\vb \triangleq \vec(\tB)$, the decomposition~\eqref{LSR_defn} can be expressed in vectorized form as 
\begin{equation*}
\bb \;=\; \sum_{s\in[S]} \paren{\B_{(K,s)}\otimes \cdots \otimes \B_{(1,s)}}\,\g.
\end{equation*}
\end{Definition}

\begin{figure}
  \centering
  \includegraphics[width=0.8\linewidth]{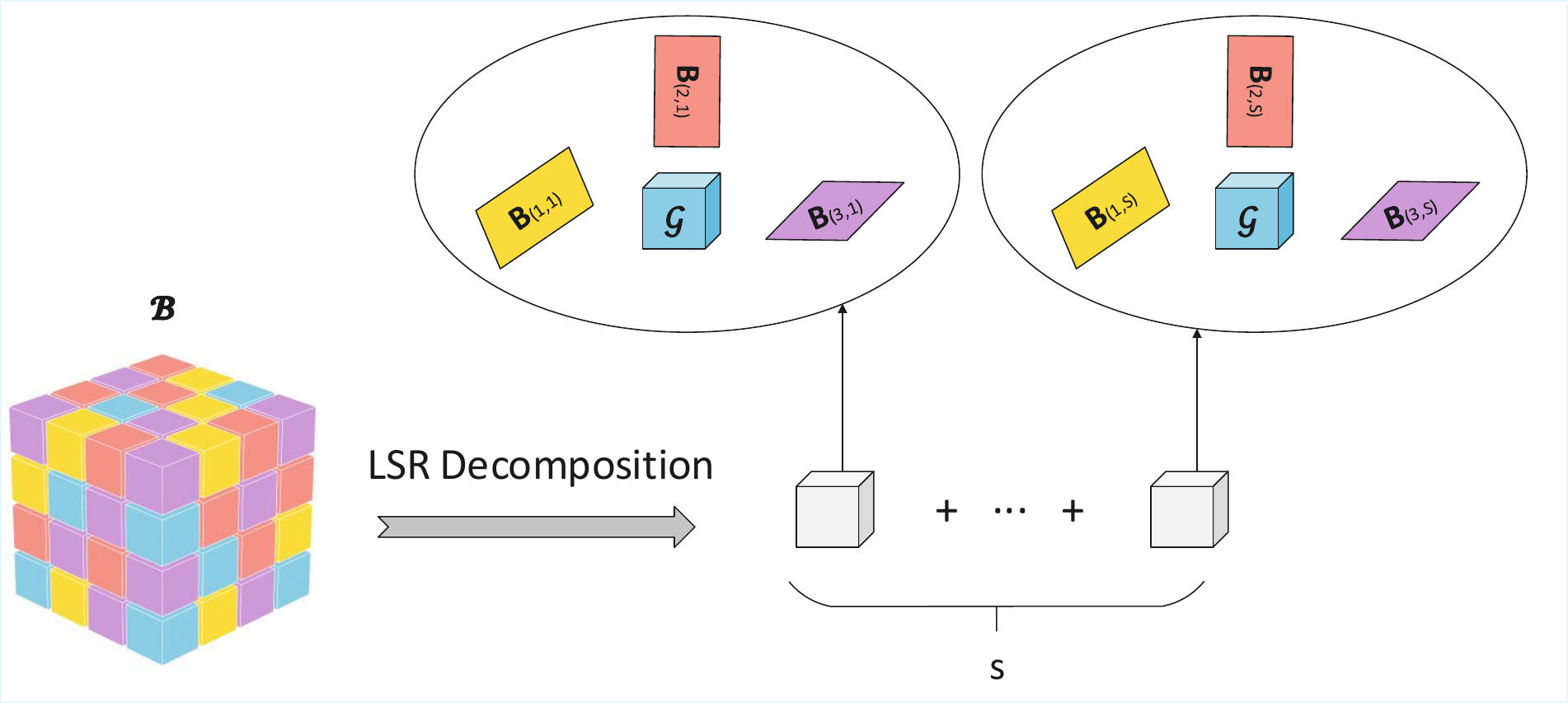}
  \caption{A third-order tensor under the LSR decomposition.}
  \label{fig:LSR_Decomposition}
\end{figure}

When $S=1$, the LSR tensor decomposition reduces to the Tucker decomposition:
\[
\tB
=
\tG \times_1 \B_{1} \times_2 \cdots \times_K \B_{K}.
\]
A schematic illustration of the LSR tensor decomposition is provided in Fig.~\ref{fig:LSR_Decomposition}. As shown in the figure, the tensor is represented as a sum of $S$ Tucker-structured components, each constructed from mode-wise factor matrices and a shared core tensor. 

\section{Tensor Generalized Linear Models}
\label{sec:Problem Statement}

Building upon the vector-structured GLMs formulation in Section~\ref{sec:Preliminaries}, 
we now extend the framework to tensor-valued covariates.

Let $\{(\tX_i, y_i)\}_{i=1}^n$ be i.i.d. training samples, where
$\tX_i \in \RR^{m_1 \times \cdots \times m_K}$ is a $K$-mode covariate tensor and
$y_i \in \RR$ is a scalar response. Conditioned on $\tX_i$, the response follows an
exponential-family distribution with natural parameter $\eta_i$ 
as defined in~\eqref{eq:exp_family_def1}. 
Under the TGLMs formulation, the natural parameter is given as $\eta_i = \langle \tB, \tX_i \rangle + z,$ where $\tB \in \RR^{m_1 \times \cdots \times m_K}$ is the coefficient tensor. For notational simplicity, we follow~\cite{taki2023structured} and set the bias term to zero, i.e., $z=0$. The conditional mean satisfies $\mu_i = \mathbb{E}[y_i \mid \tX_i] = g^{-1}(\eta_i),$ where $g(\cdot)$ is an invertible link function.

To reduce the complexity of the model and exploit the multilinear structure, 
we assume that the coefficient tensor admits an LSR decomposition 
as defined in Definition~\ref{def:LSR_tensor}. Specifically,

\begin{equation}\label{eq:lsrB}
\tB \triangleq \sum_{s=1}^{S} \tG \times_1 \mB_{(1,s)} \times_2 \cdots \times_K \mB_{(K,s)},
\end{equation}
where the core tensor $\tG \in \RR^{r_1 \times \cdots \times r_K}$ is shared across all components,
and each factor matrix $\mB_{(k,s)}$ is assumed to have orthonormal columns, 
i.e. $\mB_{(k,s)}^\top \mB_{(k,s)} = \mathbf{I}_{r_k}$.

The objective of regression is to predict the response variable from the observed covariates by estimating the underlying model parameters. Under the LSR-TGLM formulation, the coefficient tensor $\tB$ is parameterized through the core tensor $\tG$ and the factor matrices $\{\mB_{(k,s)}\}$ as defined in~\eqref{eq:lsrB}. Parameter estimation therefore amounts to learning $\tG$ and $\{\mB_{(k,s)}\}$ from the data samples  $\{(\tX_i, y_i)\}_{i=1}^n$. 
Using the exponential-family representation in~\eqref{eq:exp_family_def1}, the empirical negative log-likelihood can be written explicitly in terms of $\tG$ and $\{\mB_{(k,s)}\}$, leading to the following factorized optimization problem~\cite{taki2023structured}:
\begin{equation}\label{eq:main-problem_v2}
\begin{aligned}
\minimize_{\{\mB_{(k,s)}\},\,\tG} \quad & L_n\big(\{\mB_{(k,s)}\},\tG\big) \\
=  \frac{1}{n}&\sum_{i=1}^{n} \Bigg[ a\Bigg(\inner{\sum_{s=1}^{S} \tG \times_1 \mB_{(1,s)} \times_2\! \cdots \times_K \mB_{(K,s)},\ \tX_i}\!\!\Bigg) \\
&-T(y_i) \!\inner{\sum_{s=1}^{S} \tG \!\times_1\! \mB_{(1,s)} \!\times_2 \cdots \times_K \mB_{(K,s)},\ \tX_i} \!\Bigg] \\
\st \quad & \mB_{(k,s)}^\top \mB_{(k,s)} = \mId_{r_k}, \qquad \forall\, k\in[K],~s\in[S].
\end{aligned}
\end{equation}

Problem \eqref{eq:main-problem_v2} is nonconvex due to both the multilinear LSR parameterization and the orthogonality constraints on the factor matrices.
To solve~\eqref{eq:main-problem_v2}, the authors in~\cite{taki2023structured} proposed the LSRTR algorithm, which adopts a block coordinate descent (BCD) strategy that alternates between updating the core tensor and the factor matrices. While this approach is well aligned with the LSR structure, each factor update requires handling orthogonality constraints through explicit QR-based projection steps. These repeated projections can substantially increase the per-iteration cost and may slow down convergence, particularly in high-dimensional settings.

In practice, the runtime of LSRTR is often dominated by the factor-matrix subproblems. This computational bottleneck motivates us to improve the factor-matrix updates within the LSRTR framework. Specifically, we introduce Muon-style orthogonalized momentum updates into the factor-matrix blocks. Instead of performing explicit orthogonal projections after each gradient step, Muon updates each factor matrix using a momentum-based orthogonalized direction computed via a lightweight Newton-Schulz procedure. 
Our objective is to accelerate the solution of~\eqref{eq:main-problem_v2} while preserving the original LSR-TGLM formulation and the outer BCD structure. The resulting algorithm, termed LSRTR-M, retains the core-tensor update of LSRTR and replaces its QR-based factor-matrix updates with Muon updates. The detailed description of LSRTR-M is presented in the next section.

\section{The Proposed LSRTR-M Algorithm}
\label{sec:Algorithm design}

In this work, we propose LSRTR-M, a Muon-accelerated optimization algorithm for LSR-TGLMs.  
LSRTR-M preserves the original BCD outer loop and the core-tensor update of LSRTR~\cite{taki2023structured}, while replacing each factor-matrix update with an orthogonalized momentum step. 
The proposed algorithm is summarized in Algorithm~\ref{alg:muon-lsrtr}.


\begin{algorithm}
\caption{LSRTR-M: A Muon-Accelerated Block Coordinate Algorithm for LSR-TGLMs}\label{alg:muon-lsrtr}
\begin{algorithmic}
\STATE \textsc{Input:} training data $\{(\tX_i,y_i)\}_{i=1}^{n}$; rank $(r_1,\ldots,r_K)$; separation rank $S$;
stepsizes $\alpha$, $\alpha_{m}$; momentum $\beta$; weight decay $\lambda$.
\STATE \textsc{Initialize:} Factor matrices $\{\mB^{0}_{(k,s)}\}_{k\in[K],\,s\in[S]}$, core tensor $\tG^{0}$; momentum matrices $\{\mM^{0}_{(k,s)}\gets \mzero\}$.
\STATE \textbf{for} $t=0,\ldots,T-1$ \textbf{do}
\STATE \hspace{0.3cm}\textbf{for} $s'=1,\ldots,S$ \textbf{do}
\STATE \hspace{0.6cm}\textbf{for} $k'=1,\ldots,K$ \textbf{do}
\STATE \hspace{0.9cm}Compute block gradient $\nabla_{\mB_{(k',s')}} L_n\!\big(\{\mB^{t}_{(k,s)}\},\tG^{t}\big)$.
\STATE \hspace{0.9cm}$\mM^{t+1}_{(k',s')} \gets \beta\,\mM^{t}_{(k',s')} + \nabla_{\mB_{(k',s')}} L_n\!\big(\{\mB^{t}_{(k,s)}\},\tG^{t}\big)$.
\STATE \hspace{0.9cm}$\mQ^{t+1}_{(k',s')} \gets \mathrm{Orth}\big(\mM^{t+1}_{(k',s')}\big)$.
\STATE \hspace{0.9cm}$\mB^{t+1}_{(k',s')} \gets \mB^{t}_{(k',s')}
-\alpha_{m}\!\left(\mQ^{t+1}_{(k',s')}+\lambda\,\mB^{t}_{(k',s')}\right)$.
\STATE \hspace{0.6cm}\textbf{end for}
\STATE \hspace{0.3cm}\textbf{end for}
\STATE \hspace{0.3cm}Compute $\nabla_{\tG} L_n\!\big(\{\mB^{t+1}_{(k,s)}\},\tG^{t}\big)$.
\STATE \hspace{0.3cm}$\tG^{t+1}\gets \tG^{t}-\alpha\,\nabla_{\tG} L_n\!\big(\{\mB^{t+1}_{(k,s)}\},\tG^{t}\big)$.
\STATE Exit if the stopping criterion is met.
\STATE \textbf{end for}
\STATE \textsc{Return:} $\widehat{\tB} \gets \sum_{s=1}^{S} \tG^{t_{\text{final}}}\times_{1}\mB^{t_{\text{final}}}_{(1,s)}\times_{2}\cdots\times_{K}\mB^{t_{\text{final}}}_{(K,s)}$.
\end{algorithmic}
\end{algorithm}


At each iteration, we first update all factor matrices $\{\B_{(k,s)}\}_{k\in[K],\,s\in[S]}$, followed by the core tensor $\tG$ update.

\paragraph{Factor matrices update}
At iteration $t$, for each block $(k',s')$, we compute the partial gradient
$\nabla_{\mB_{(k',s')}} L_n\!\paren{\{\mB^{t}_{(k,s)}\},\tG^{t}}$ and update the corresponding momentum accumulator by
\begin{equation*}
\mM^{t+1}_{(k',s')}
\;=\;
\beta\,\mM^{t}_{(k',s')}
\;+\;
\nabla_{\mB_{(k',s')}} L_n\!\big(\{\mB^{t}_{(k,s)}\},\tG^{t}\big).
\label{eq:muon_momentum}
\end{equation*}

We then construct the update direction via the orthogonalization operator:
\begin{equation*}
\begin{aligned}
\mQ^{t+1}_{(k',s')}
&\;=\;
\mathrm{Orth}\paren{\mM^{t+1}_{(k',s')}}\;\triangleq\;
\mM^{t+1}_{(k',s')}\!\Big(\!\big(\mM^{t+1}_{(k',s')}\big)\!^{\top}\mM^{t+1}_{(k',s')}\Big)^{-1/2},
\label{eq:right_orth}
\end{aligned}
\end{equation*}
which yields $\big(\mQ^{t+1}_{(k',s')}\big)^{\top}\mQ^{t+1}_{(k',s')} = \mId$ when the inverse square root is computed exactly.
The factor matrix is then updated using a gradient-like step with weight decay:
\begin{equation*}
\mB^{t+1}_{(k',s')}
\;=\;
\mB^{t}_{(k',s')}
\;-\;
\alpha_{m}\Big(\mQ^{t+1}_{(k',s')}+\lambda\,\mB^{t}_{(k',s')}\Big),
\label{eq:muon_factor_update}
\end{equation*}
where $\lambda$ is a weight-decay parameter.

Unlike LSRTR, which performs a Euclidean gradient step followed by an explicit projection onto the orthogonality constraint (e.g., via QR), LSRTR-M integrates orthogonalization directly into the update direction. Specifically, the accumulated momentum is normalized through a polar-type transformation before applying the step. 
This avoids repeatedly correcting the iterate after an unconstrained update and leads to more efficient search directions.
In addition, the inverse square root of the small $r_k \times r_k$ Gram matrix can be efficiently approximated via a few Newton-Schulz iterations~\cite{higham1986computing}. These effects lead to more stable optimization trajectories and improved empirical performance, as demonstrated in our experiments.

\paragraph{Core tensor update}
After updating all factor blocks, we update the core tensor via a standard gradient descent step:
\begin{equation*}
\tG^{t+1}\gets \tG^{t}-\alpha\,\nabla_{\tG} L_n\!\big(\{\mB^{t+1}_{(k,s)}\},\tG^{t}\big)
\label{eq:core_update}
\end{equation*}
Unlike the factor-matrix updates, the core tensor update does not involve orthogonality constraints and therefore remains unchanged from the original LSRTR framework.
Upon termination, the final estimate $\widehat{\tB}$ 
is reconstructed from the last iterations 
$\tG^{t_{\mathrm{final}}}$ and $\{\B^{t_{\mathrm{final}}}_{(k,s)}\}$ 
according to the LSR decomposition~\eqref{LSR_defn}.

\section{Experiments}
\label{sec:Experiment}

In this section, we conduct comprehensive numerical experiments to evaluate the proposed LSRTR-M algorithm against the baseline LSRTR method~\cite{taki2023structured} for LSR-TGLMs. Our study focuses on three aspects: 1) Optimization behavior: convergence speed across iterations and wall-clock time; 2) Statistical performance: estimation and prediction accuracy under varying sample sizes; 3) Real-data evaluation: classification performance on the Vessel MNIST 3D medical imaging dataset~\cite{yang2020intra,yang2021medmnist}.
All experiments were implemented in MATLAB R2024a and conducted on a server running Red Hat Enterprise Linux 9.3, equipped with an Intel(R) Xeon(R) Gold 6542Y processor and 502 GB RAM.

\subsection{Synthetic Data}

We first evaluate the two algorithms on synthetic datasets generated under the three canonical TGLM settings: linear (Gaussian), logistic (Bernoulli), and Poisson regression. The performance is assessed using: (i) the training loss, (ii) the normalized estimation error $\frac{\left\|\tB-\widehat{\tB}\right\|_{F}^{2}}{\left\|\tB\right\|_{F}^{2}}$, and (iii) prediction error measured on an independent test set. 
Let $n_{te}$ denote the number of test samples, 
$\vy \in \RR^{n_{te}}$ the vector of ground-truth responses, 
and $\hat{\vy} \in \RR^{n_{te}}$ the corresponding predicted responses.
For linear regression, prediction error is measured by the normalized squared error $\frac{\|\hat{\vy}-\vy\|_2^2}{\|\vy\|_2^2}$. For logistic regression, we report the mean absolute error (MAE) $\frac{\|\hat{\vy}-\vy\|_1}{n_{te}}$ for the $n_{te}$ testing samples. For Poisson regression, we use the normalized squared logarithmic error $\frac{\|\log(\hat{\vy}+1)-\log(\vy+1)\|_2^2}{\|\log(\vy+1)\|_2^2}$. Unless otherwise specified, we set the tensor order to $K=3$, the separation rank to $S=2$, the multilinear rank to $\r=(r_1,\ldots,r_K)= (2,2,2)$, and the tensor dimension to
$m=(10,15,20)$. All results are averaged over 50 independent trials.

\subsubsection{Linear Regression (Gaussian)}


In the linear regression setting, the LSR-TGLMs problem~\eqref{eq:main-problem_v2} reduces to the following factorized problem
\begin{equation*}\label{eq:obj_linear_main}
\begin{aligned}
&\min_{\{\mB_{(k,s)}\},\,\tG}\quad
\mathcal{L}_{\mathrm{lin}}\big(\{\mB_{(k,s)}\},\tG\big) \\
\triangleq\;& \frac{1}{2n}\sum_{i=1}^{n}
\Bigg(
y_i-
\inner{
\sum_{s=1}^{S}\tG \times_1 \mB_{(1,s)} \times_2 \cdots \times_K \mB_{(K,s)},
\ \tX_i
}
\Bigg)^2 \\
&\st\quad
 \mB_{(k,s)}^\top \mB_{(k,s)}=\mId_{r_k},
\qquad \forall\,k\in[K],\ s\in[S],
\end{aligned}
\end{equation*}
which is solved by LSRTR and LSRTR-M with different factor-matrix update rules.

\begin{figure}[t]
\centering

\begin{subfigure}{0.32\linewidth}
\centering
\includegraphics[width=\linewidth]{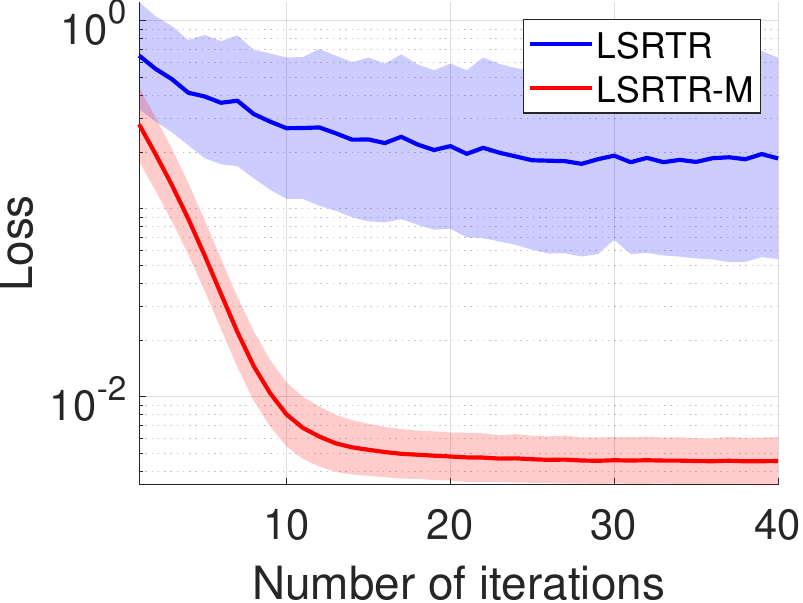}
\caption{}
\label{fig:loss_iter}
\end{subfigure}
\hfill
\begin{subfigure}{0.32\linewidth}
\centering
\includegraphics[width=\linewidth]{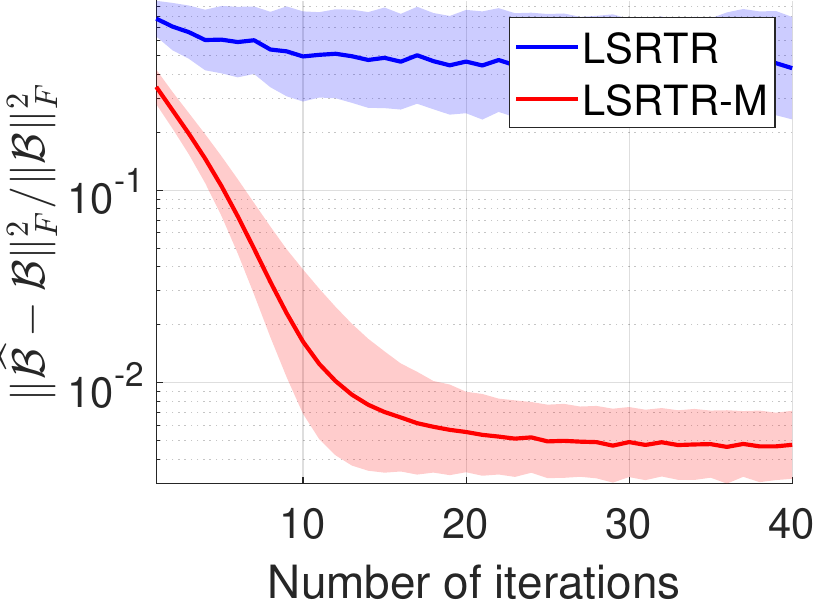}
\caption{}
\label{fig:est_iter}
\end{subfigure}
\hfill
\begin{subfigure}{0.32\linewidth}
\centering
\includegraphics[width=\linewidth]{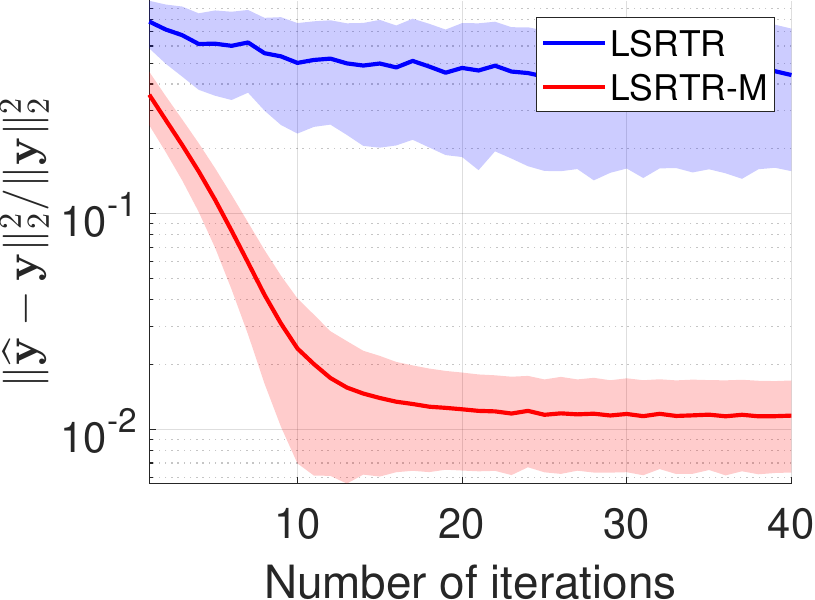}
\caption{}
\label{fig:pred_iter}
\end{subfigure}

\vspace{0.5em}

\begin{subfigure}{0.32\linewidth}
\centering
\includegraphics[width=\linewidth]{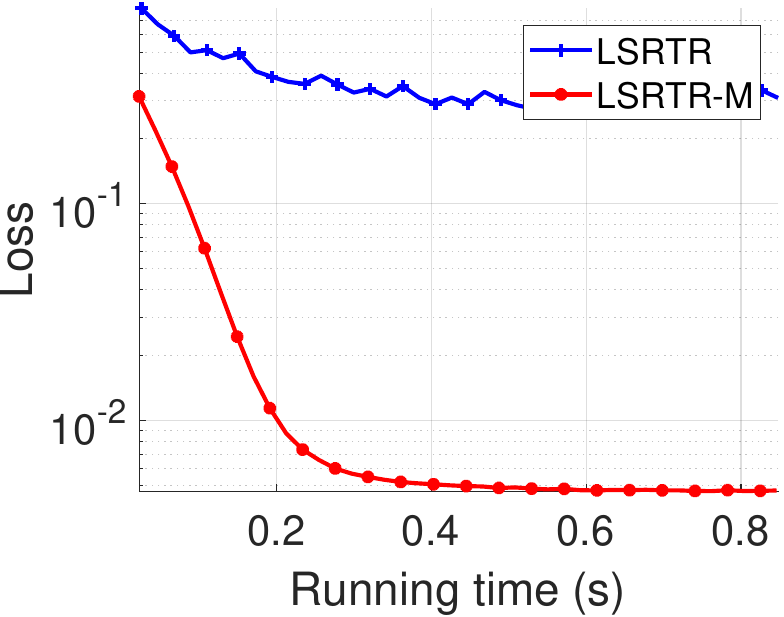}
\caption{}
\label{fig:loss_time}
\end{subfigure}
\hfill
\begin{subfigure}{0.32\linewidth}
\centering
\includegraphics[width=\linewidth]{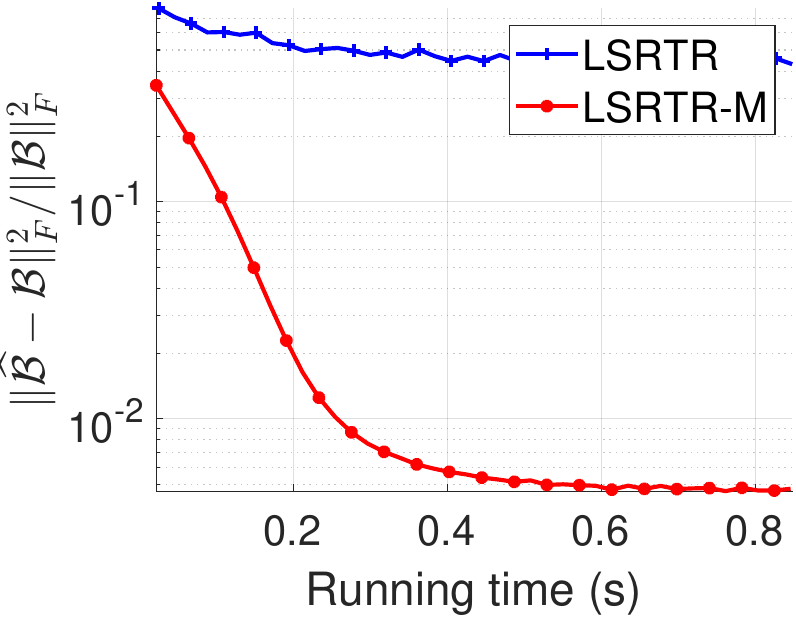}
\caption{}
\label{fig:est_time}
\end{subfigure}
\hfill
\begin{subfigure}{0.32\linewidth}
\centering
\includegraphics[width=\linewidth]{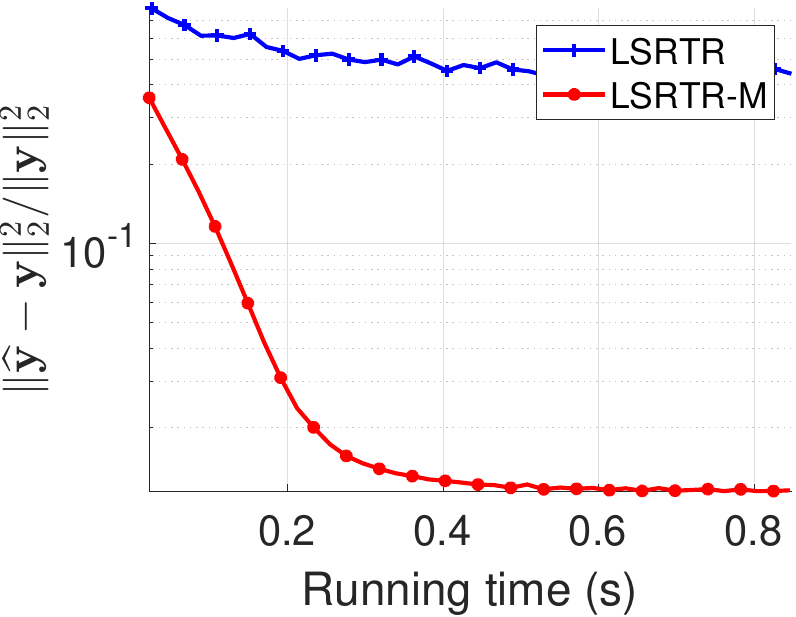}
\caption{}
\label{fig:pred_time}
\end{subfigure}

\caption{Performance comparison in linear regression. Top row: results versus iterations. Bottom row: results versus running time. Columns correspond to training loss, normalized estimation error, and normalized prediction error, respectively.}
\label{fig:linear_combined}
\end{figure}

We generate $n=500$  training samples $\{(\tX_i, y_i)\}_{i=1}^n$ and $n_{te}=100$  independent test samples for prediction evaluation. The response variable is generated according to $y_i \sim \mathcal{N}(\langle \tB, \tX_i \rangle,\, 0.1)$. Both algorithms are initialized in a neighborhood of the ground truth. Specifically, the core tensor is initialized by adding small Gaussian perturbations to the true core, and each factor matrix is initialized by perturbing the corresponding ground-truth factor with Gaussian noise followed by orthonormalization. For LSRTR-M, all momentum matrices are initialized to zero. Both algorithms are executed for 40 iterations. 
LSRTR uses a fixed stepsize $\alpha=0.5$, whereas LSRTR-M adopts 
$\alpha_m=0.05$ with momentum parameter $\beta=0.1$ and weight decay $\lambda = 10^{-3}$.
As shown in Figure~\ref{fig:linear_combined}, LSRTR-M decreases the training loss, normalized estimation error, and prediction error significantly faster than LSRTR. Within the same number of iterations, LSRTR-M attains substantially lower error levels.  
The same trend is observed when performance is measured against wall-clock time, where LSRTR-M reaches a given accuracy level much earlier. These results demonstrate that LSRTR-M improves both convergence speed and practical computational efficiency.
In Figure~\ref{fig:linear_combined} (top row), the shaded regions indicate one standard deviation around the mean over repeated trials. The relatively narrower shaded regions of LSRTR-M indicate lower variability across runs and suggest more stable empirical performance.

\begin{figure}
\centering
\begin{subfigure}{0.32\linewidth}
\centering
\includegraphics[width=\linewidth]{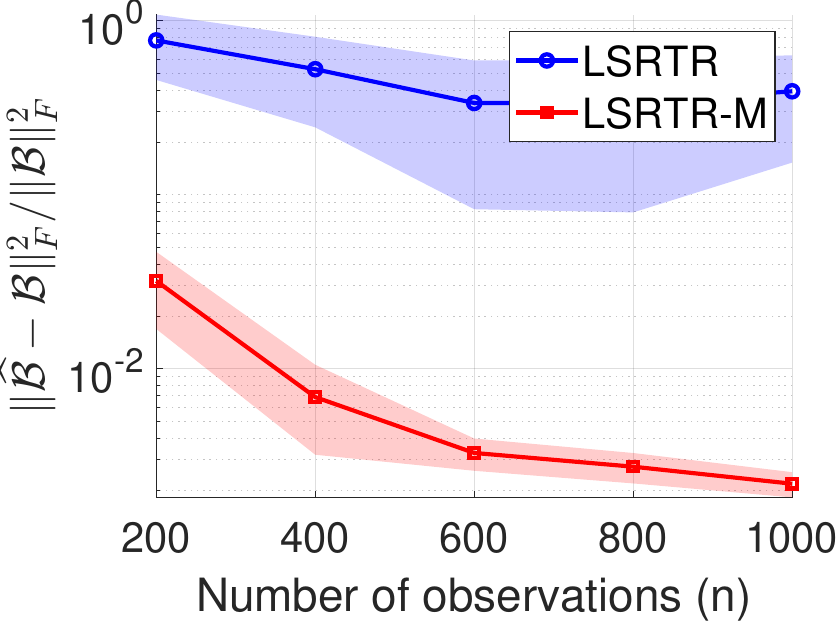}
\caption{}
\label{fig:est_iter}
\end{subfigure}
\hfill
\begin{subfigure}{0.32\linewidth}
\centering
\includegraphics[width=\linewidth]{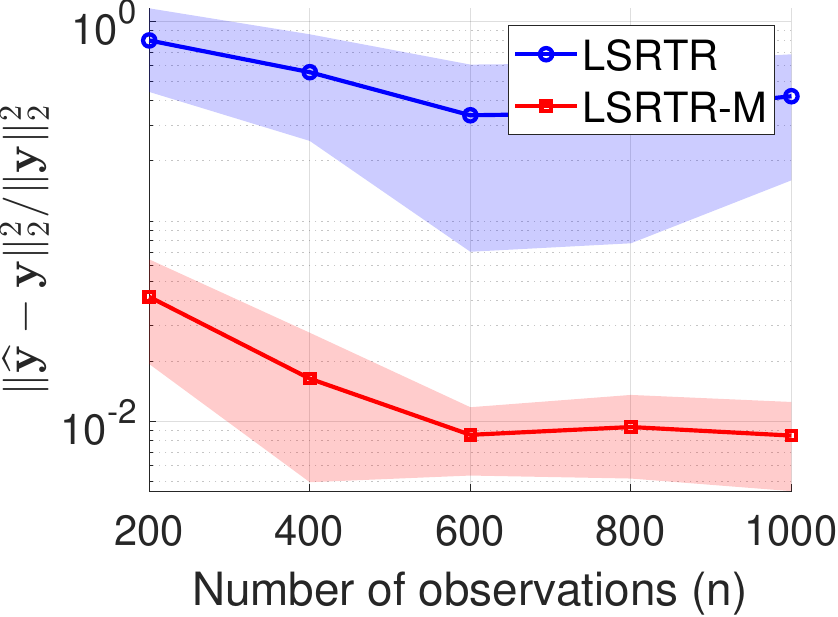}
\caption{}
\label{fig:pred_iter}
\end{subfigure}
\caption{
Performance comparison across training sample sizes in linear regression. (a) Normalized estimation error and (b) normalized prediction error.
}
\label{fig:observations_linear}
\end{figure}

To assess the effect of sample size, we additionally vary the number of training observations $n$ while fixing the number of test samples at $n_{te}=100$, and use the same problem dimensions, initialization, and algorithmic parameters as in the previous experiment. Figure~\ref{fig:observations_linear} reports the mean normalized estimation and prediction errors, with shaded regions indicating one standard deviation over repeated trials. As $n$ increases, both methods benefit from additional data. However, LSRTR-M consistently achieves substantially lower estimation and prediction errors across all tested sample sizes. Moreover, the smaller variance bands indicate improved stability of LSRTR-M relative to LSRTR. These results demonstrate that the proposed Muon-based updates provide not only faster optimization but also more accurate and reliable parameter recovery as the data scale grows.

\subsubsection{Logistic Regression (Bernoulli)}


In the logistic regression setting, the LSR-TGLMs Problem~\eqref{eq:main-problem_v2} becomes
\begin{equation*}\label{eq:loss_logistic}
\begin{aligned}
&\min_{\{\mB_{(k,s)}\},\,\tG}\quad
\mathcal{L}_{\mathrm{logit}}\big(\{\mB_{(k,s)}\},\tG\big) \\
\triangleq\;
& \frac{1}{n}\sum_{i=1}^{n}
\Bigg[
\log\Bigg(\!\!1\!+\!\exp\!\Big(
\inner{
\sum_{s=1}^{S}\tG \times_1 \mB_{(1,s)} \times_2 \cdots \times_K \mB_{(K,s)},
 \tX_i
}
\Big)\Bigg) \\
&- y_i\,
\inner{
\sum_{s=1}^{S}\tG \times_1 \mB_{(1,s)} \times_2 \cdots \times_K \mB_{(K,s)},
\ \tX_i
}
\Bigg] \\
&\st\quad
\mB_{(k,s)}^\top \mB_{(k,s)}=\mId_{r_k},
\qquad \forall\,k\in[K],\ s\in[S].
\end{aligned}
\end{equation*}

\begin{figure}[t]
\centering

\begin{subfigure}{0.32\linewidth}
\centering
\includegraphics[width=\linewidth]{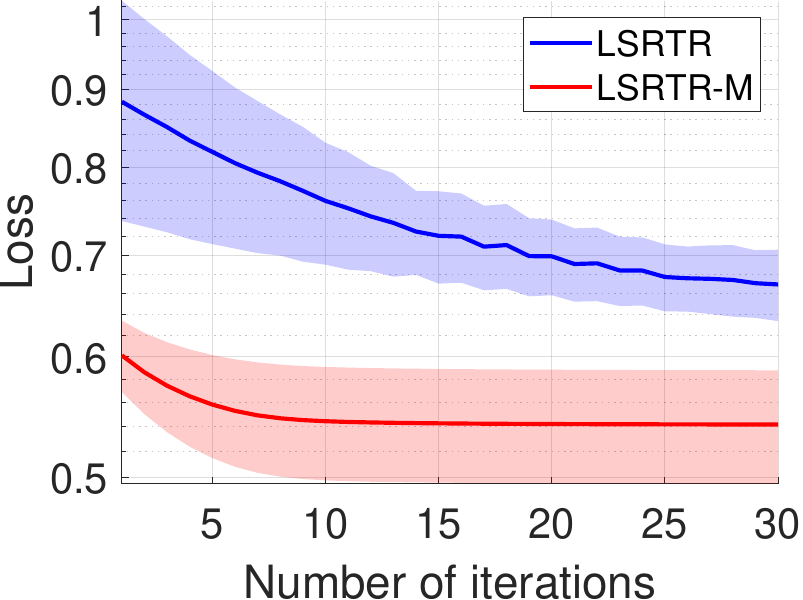}
\caption{}
\label{fig:loss_iter_logistic}
\end{subfigure}
\hfill
\begin{subfigure}{0.32\linewidth}
\centering
\includegraphics[width=\linewidth]{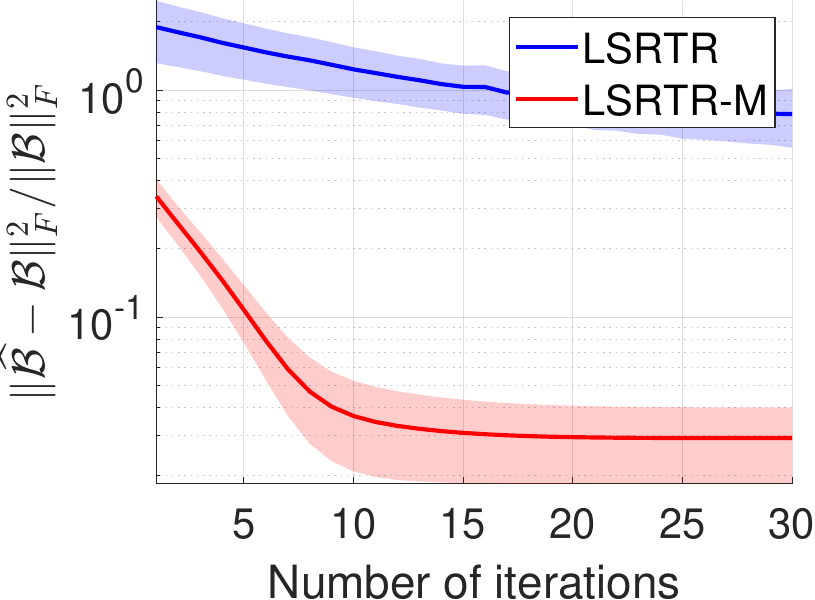}
\caption{}
\label{fig:est_iter_logistic}
\end{subfigure}
\hfill
\begin{subfigure}{0.32\linewidth}
\centering
\includegraphics[width=\linewidth]{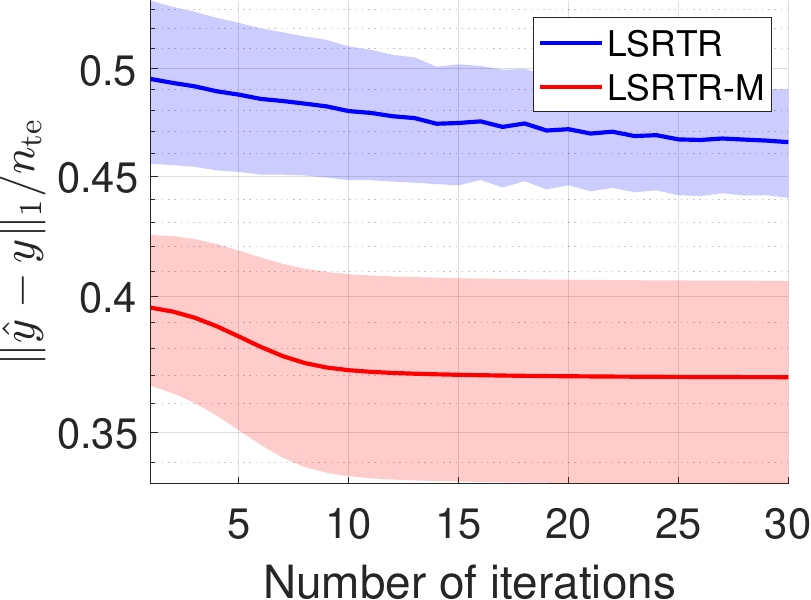}
\caption{}
\label{fig:pred_iter_logistic}
\end{subfigure}

\vspace{0.5em}

\begin{subfigure}{0.32\linewidth}
\centering
\includegraphics[width=\linewidth]{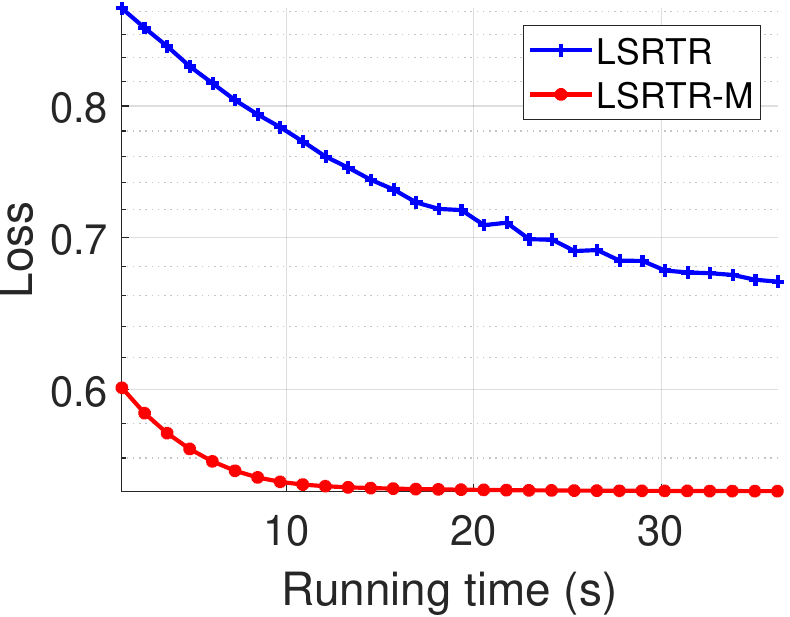}
\caption{}
\label{fig:loss_time_logistic}
\end{subfigure}
\hfill
\begin{subfigure}{0.32\linewidth}
\centering
\includegraphics[width=\linewidth]{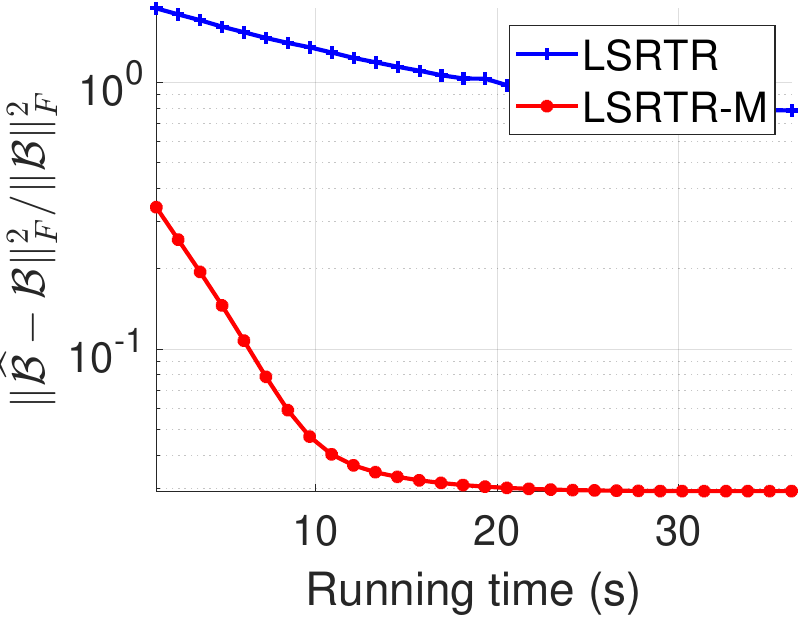}
\caption{}
\label{fig:est_time_logistic}
\end{subfigure}
\hfill
\begin{subfigure}{0.32\linewidth}
\centering
\includegraphics[width=\linewidth]{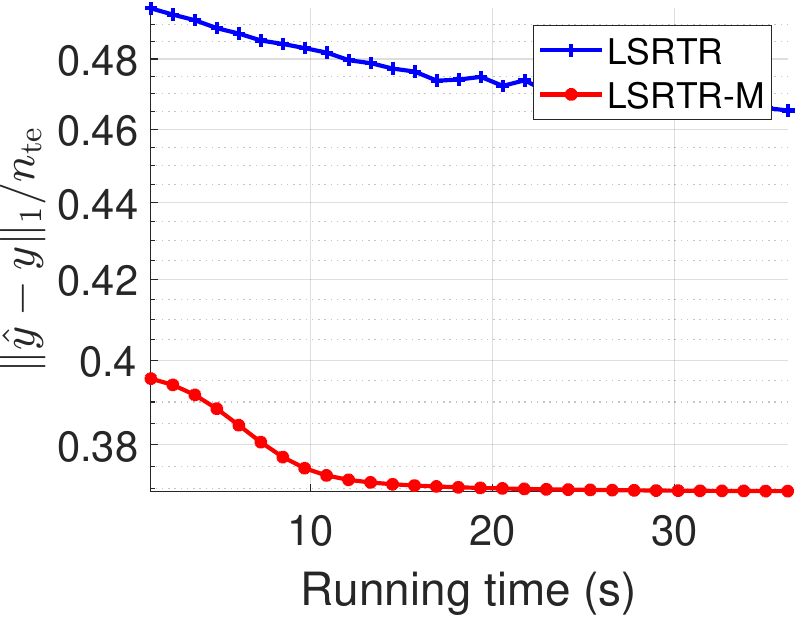}
\caption{}
\label{fig:pred_time_logistic}
\end{subfigure}

\caption{Performance comparison in logistic regression. Top row: results versus iterations. Bottom row: results versus running time. Columns correspond to training loss, normalized estimation error, and normalized prediction error, respectively.}
\label{fig:logistic_combined}
\end{figure}

We generate $n=20000$ training samples and $n_{te}=10000$ independent test samples. The response variable follows the Bernoulli model $y_i \sim \mathrm{Bernoulli}\!\left(\frac{1}{1+\exp(-\langle \tB,\tX_i\rangle)}\right)$. We use the same initialization method as in linear regression. Both LSRTR and LSRTR-M are run for $30$ iterations. LSRTR uses a fixed stepsize $\alpha=0.1$, whereas LSRTR-M adopts $\alpha_m=0.05$ with a momentum parameter $\beta=0.1$ and weight-decay $\lambda=10^{-3}$.
As shown in Figure~\ref{fig:logistic_combined}, LSRTR-M exhibits a consistently faster decrease in training loss, normalized estimation error, and prediction error, both across iterations and in terms of wall-clock time.

\begin{figure}
\centering

\begin{subfigure}{0.32\linewidth}
\centering
\includegraphics[width=\linewidth]{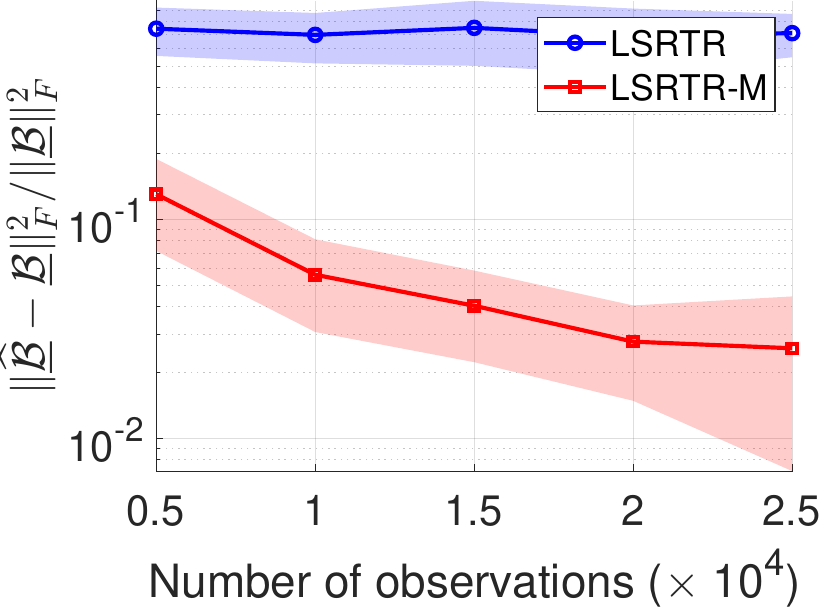}
\caption{}
\label{fig:pred_iter}
\end{subfigure}
\hfill
\begin{subfigure}{0.32\linewidth}
\centering
\includegraphics[width=\linewidth]{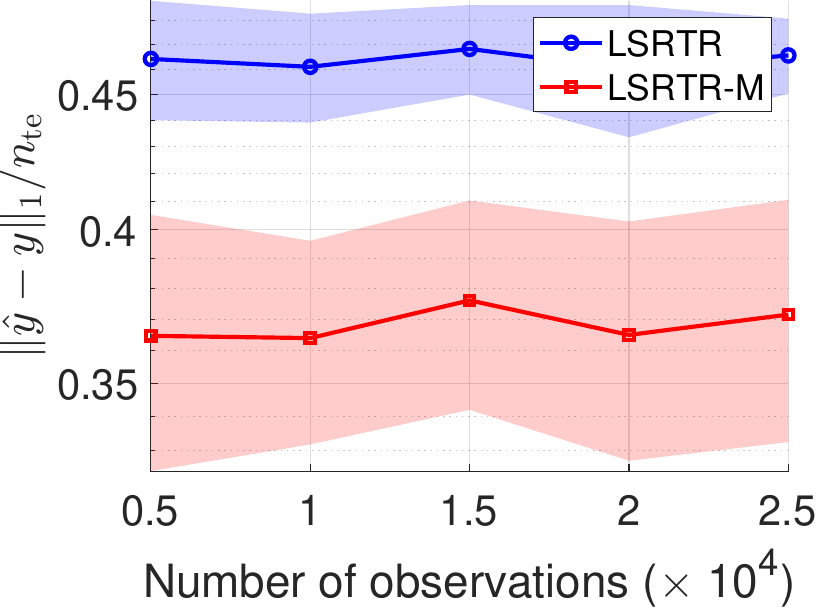}
\caption{}
\label{fig:grad_iter}
\end{subfigure}

\caption{Performance comparison across training sample sizes in logistic regression. (a) Normalized estimation error and (b) normalized prediction error.}
\label{fig:observation_logistic}
\end{figure}

Next, we fix the number of test samples at $n_{te}=5000$ and vary the number of training observations $n$ to assess the effect of sample size. Under this setting, the LSRTR stepsize is adjusted to $\alpha=0.5$, while all other hyperparameters remain unchanged. Figure~\ref{fig:observation_logistic} reports the normalized estimation and prediction errors, with shaded regions indicating one standard deviation of the mean normalized errors. Across all tested values of $n$, LSRTR-M consistently achieves substantially lower estimation and prediction errors than LSRTR. The shaded regions indicate comparable variability across trials, while maintaining a clear performance gap in favor of LSRTR-M.





\subsubsection{Poisson Regression}

In the Poisson regression setting, the LSR-TGLMs Problem~\eqref{eq:main-problem_v2} specializes to
\begin{equation*}\label{eq:loss_pois}
\begin{aligned}
&\min_{\{\mB_{(k,s)}\},\,\tG}\quad
\mathcal{L}_{\mathrm{pois}}\big(\{\mB_{(k,s)}\},\tG\big) \\
\triangleq\;
& \frac{1}{n}\sum_{i=1}^{n}
\Bigg[
\exp\!\Big(
\inner{
\sum_{s=1}^{S}\tG \times_1 \mB_{(1,s)} \times_2 \cdots \times_K \mB_{(K,s)},
\ \tX_i
}
\Big) \\
&- y_i\,
\inner{
\sum_{s=1}^{S}\tG \times_1 \mB_{(1,s)} \times_2 \cdots \times_K \mB_{(K,s)},
\ \tX_i
}
\Bigg] \\
&\st\quad
 \mB_{(k,s)}^\top \mB_{(k,s)}=\mId_{r_k},
\qquad \forall\,k\in[K],\ s\in[S].
\end{aligned}
\end{equation*}

\begin{figure}[t]
\centering

\begin{subfigure}{0.32\linewidth}
\centering
\includegraphics[width=\linewidth]{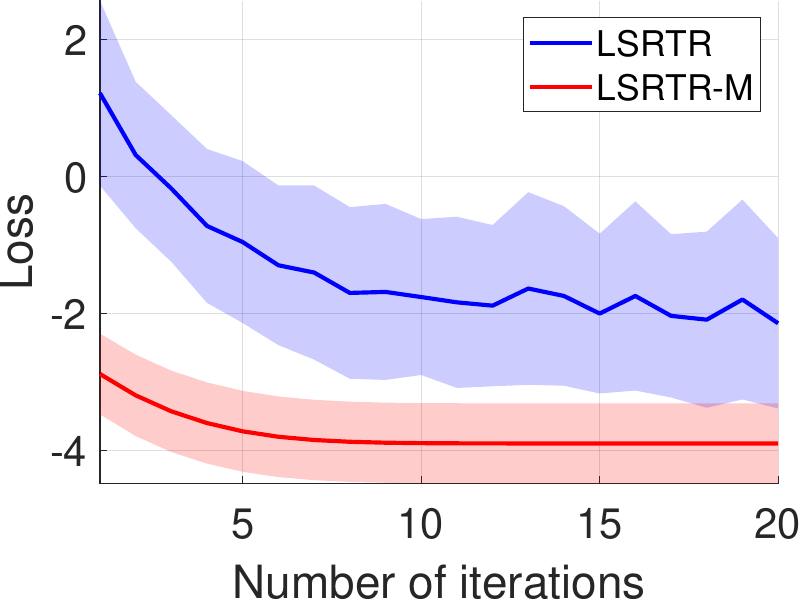}
\caption{}
\label{fig:loss_iter_poisson}
\end{subfigure}
\hfill
\begin{subfigure}{0.32\linewidth}
\centering
\includegraphics[width=\linewidth]{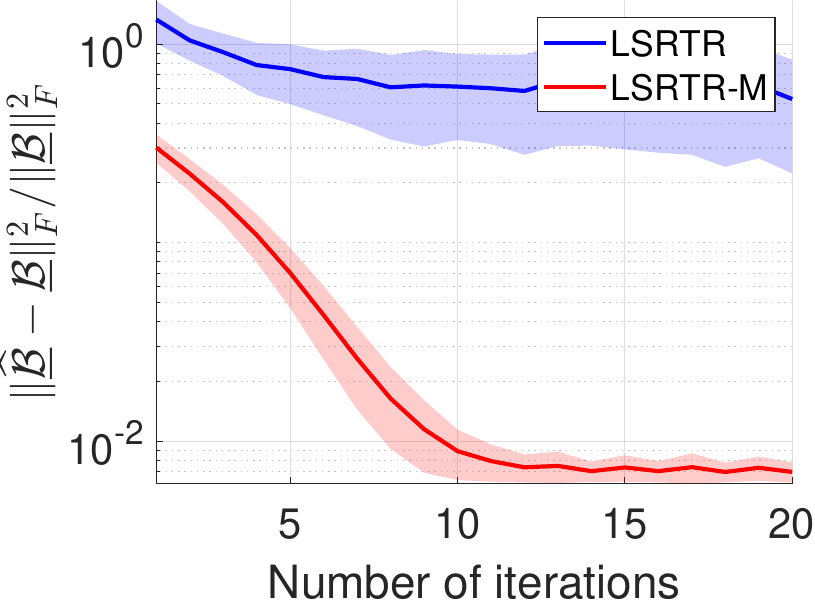}
\caption{}
\label{fig:est_iter_poisson}
\end{subfigure}
\hfill
\begin{subfigure}{0.32\linewidth}
\centering
\includegraphics[width=\linewidth]{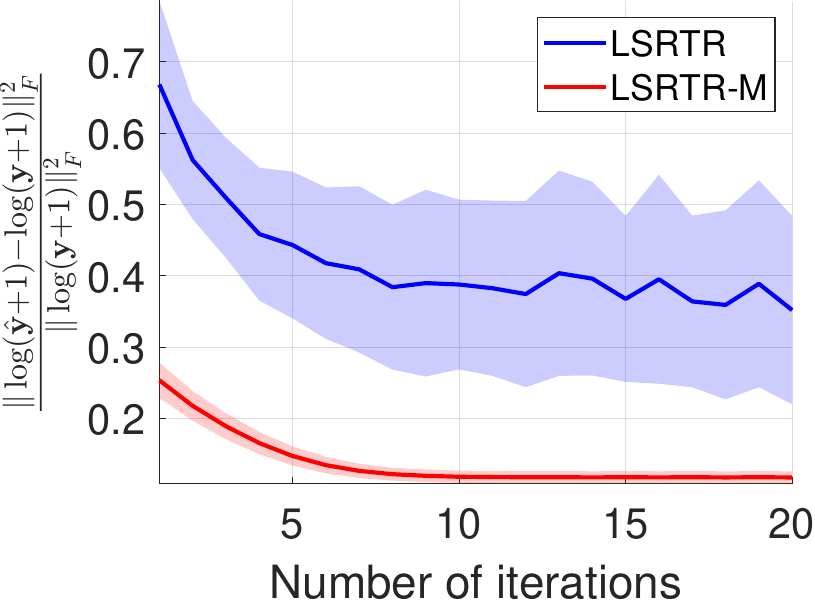}
\caption{}
\label{fig:pred_iter_poisson}
\end{subfigure}

\vspace{0.5em}

\begin{subfigure}{0.32\linewidth}
\centering
\includegraphics[width=\linewidth]{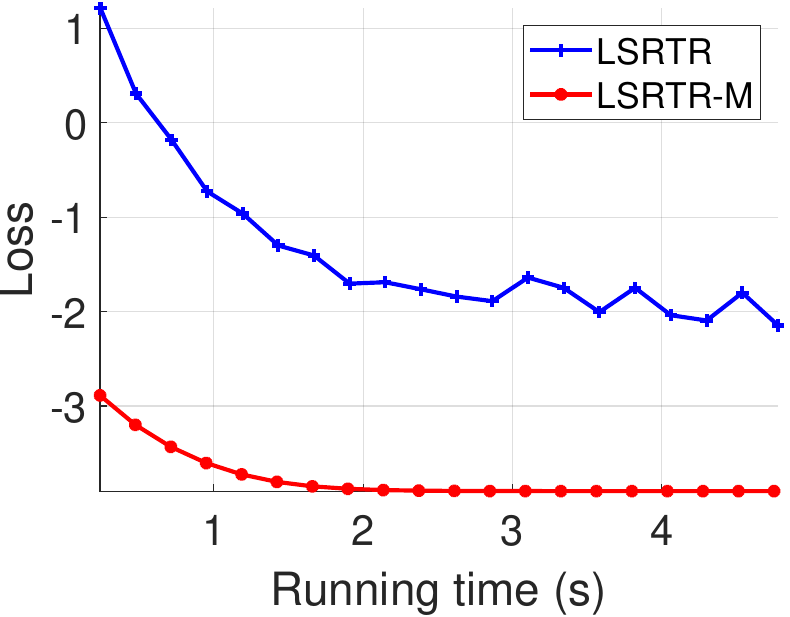}
\caption{}
\label{fig:loss_time_poisson}
\end{subfigure}
\hfill
\begin{subfigure}{0.32\linewidth}
\centering
\includegraphics[width=\linewidth]{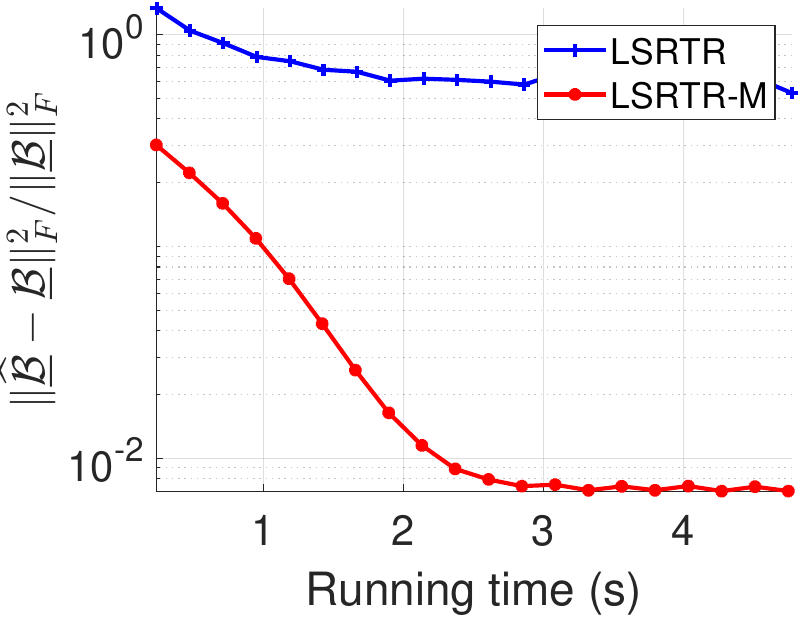}
\caption{}
\label{fig:est_time_poisson}
\end{subfigure}
\hfill
\begin{subfigure}{0.32\linewidth}
\centering
\includegraphics[width=\linewidth]{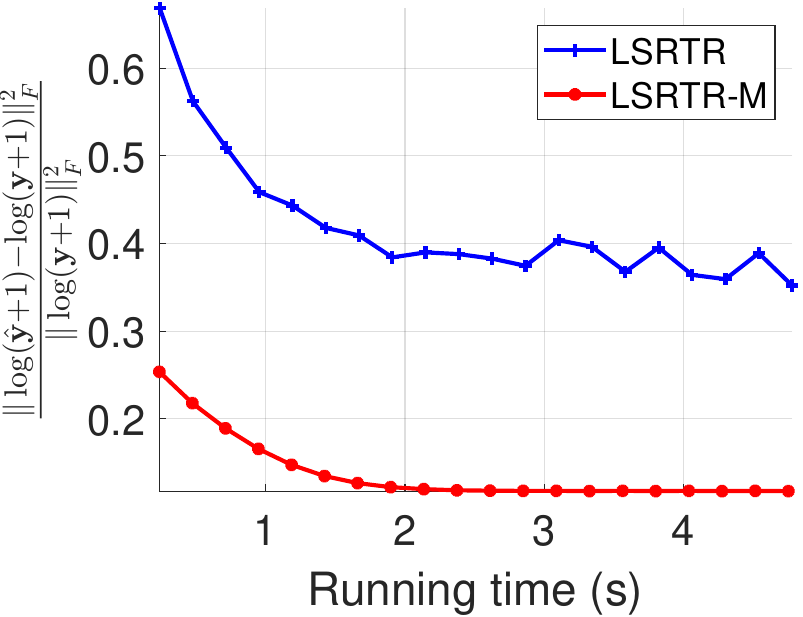}
\caption{}
\label{fig:pred_time_poisson}
\end{subfigure}

\caption{Performance comparison in Poisson regression. Top row: results versus iterations. Bottom row: results versus running time. Columns correspond to training loss, normalized estimation error, and normalized prediction error, respectively.}
\label{fig:poisson_combined}
\end{figure}

We generate $n=5000$ training samples and $n_{te}=1000$ test samples. The observation $y_i$ conditioned on data $\tX_i$ follows a Poisson distribution 
$
y_i \sim \mathrm{Poisson}\!\Big(\exp\!\big(\inner{\tB,\tX_i}\big)\Big).
$
Both methods are run for $20$ iterations with identical step size $0.05$, and LSRTR-M additionally uses a momentum parameter $\beta=0.1$ and weight-decay $\lambda=10^{-3}$. We use the same initialization method as in linear regression. Figure~\ref{fig:poisson_combined}  reports convergence behavior across iterations and running time. As in the linear and logistic settings, LSRTR-M consistently outperforms LSRTR in terms of convergence speed and final estimation and prediction accuracy.

\begin{figure}
\centering

\begin{subfigure}{0.32\linewidth}
\centering
\includegraphics[width=\linewidth]{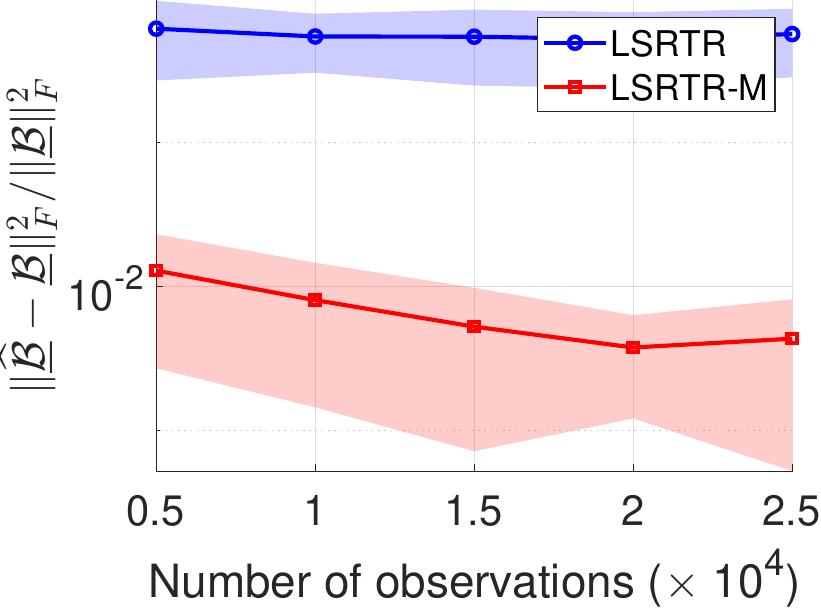}
\caption{}
\label{fig:errB_observation_Poisson}
\end{subfigure}
\hfill
\begin{subfigure}{0.32\linewidth}
\centering
\includegraphics[width=\linewidth]{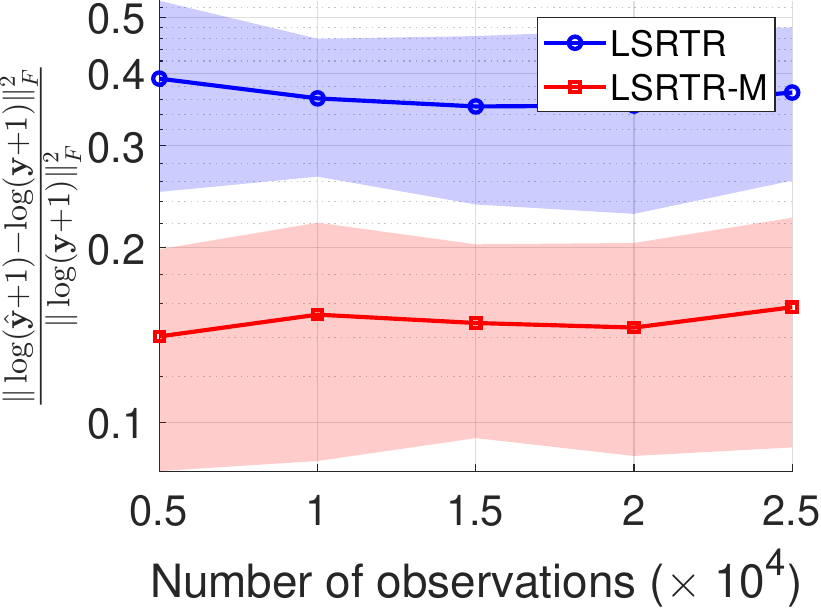}
\caption{}
\label{fig:erry_observation_Poisson}
\end{subfigure}
\hfill
\begin{subfigure}{0.32\linewidth}
\centering
\includegraphics[width=\linewidth]{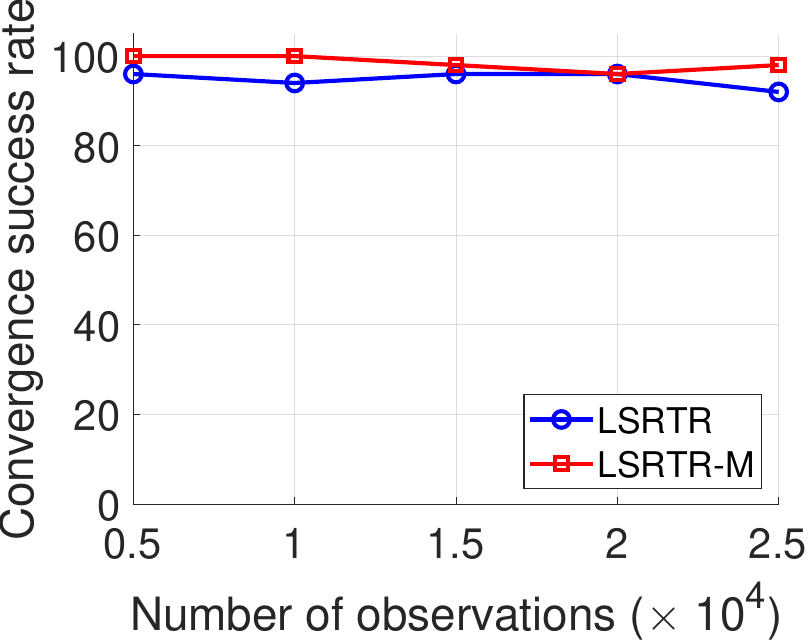}
\caption{}
\label{fig:observation_rate_Poisson}
\end{subfigure}

\caption{Performance and convergence comparison across training sample sizes in Poisson regression. (a) Normalized estimation error, (b) normalized squared logarithmic prediction error, and (c) convergence success rate ($\%$ of converged trials). }
\label{fig:observation_Poisson}
\end{figure}

To investigate scalability, we again vary the number of training samples while fixing $n_{te}=5000$. Convergence statistics are computed after excluding non-convergent trials (e.g., runs that produce NaN values). As shown in Figure~\ref{fig:observation_rate_Poisson}, LSRTR-M achieves a consistently higher convergence success rate across all considered sample sizes. Furthermore, among the converged trials, Figures~\ref{fig:errB_observation_Poisson} and~\ref{fig:erry_observation_Poisson} indicate that LSRTR-M attains uniformly lower normalized estimation and prediction errors over the entire range of $n$.

\subsection{Real Data: VESSEL MNIST 3D}

\begin{figure}[!t]
\centering
\includegraphics[width=0.50\linewidth]{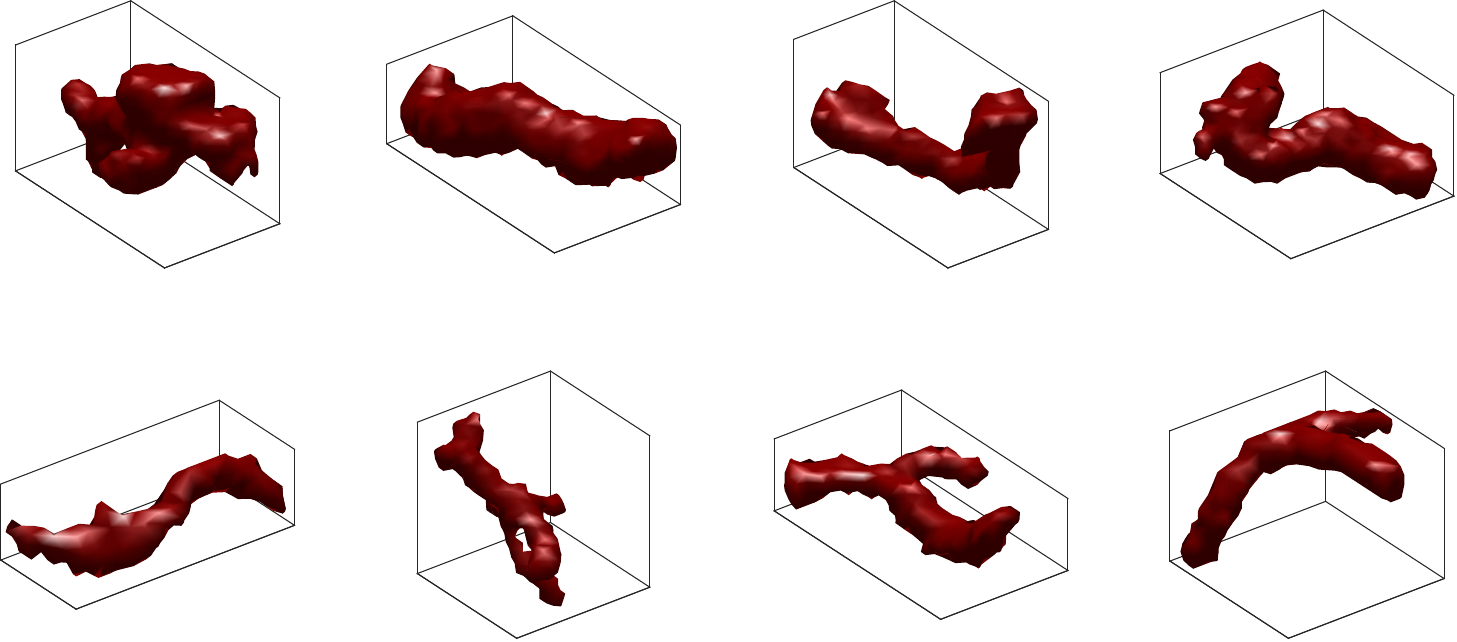}
\caption{Representative $28 \times 28 \times 28$ vessel volumes from the Vessel MNIST 3D dataset. Top row: aneurysm samples ($y=1$). Bottom row: healthy samples ($y=0$). }
\label{fig:vessel_examples}
\end{figure}

We further evaluate the two algorithms LSRTR and LSRTR-M on the Vessel MNIST 3D dataset~\cite{yang2021medmnist,yang2020intra, yang2023medmnist}, which consists of $28\times 28\times 28$ 3D vessel volumes with binary labels indicating aneurysm ($y=1$) or healthy ($y=0$). 
The dataset contains 1909 vessel segments collected from multiple subjects and has been preprocessed to remove incomplete and duplicated scans. We adopt the predefined train-test split and formulate the task as a binary logistic regression with third-order tensor covariates. Figure~\ref{fig:vessel_examples} visualizes representative samples to illustrate the volumetric structure of the covariates.

\begin{figure}[t]
  \centering
  \includegraphics[width=0.48\linewidth]{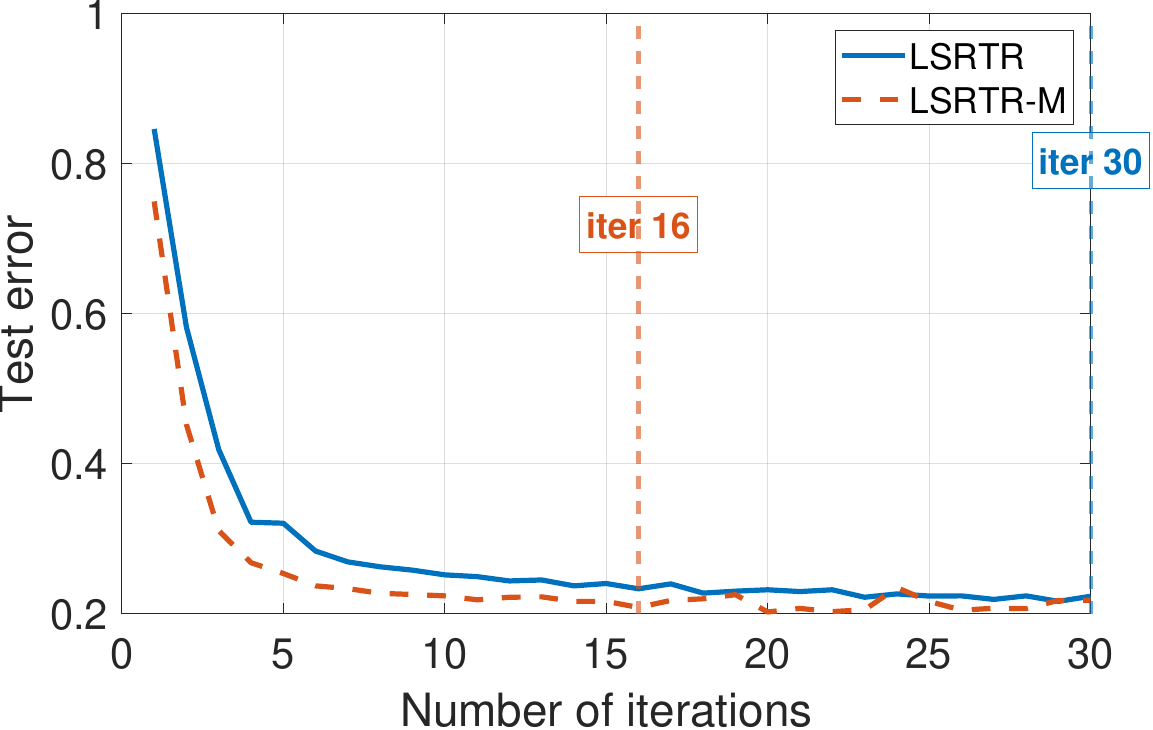}
  \hfill
  \includegraphics[width=0.48\linewidth]{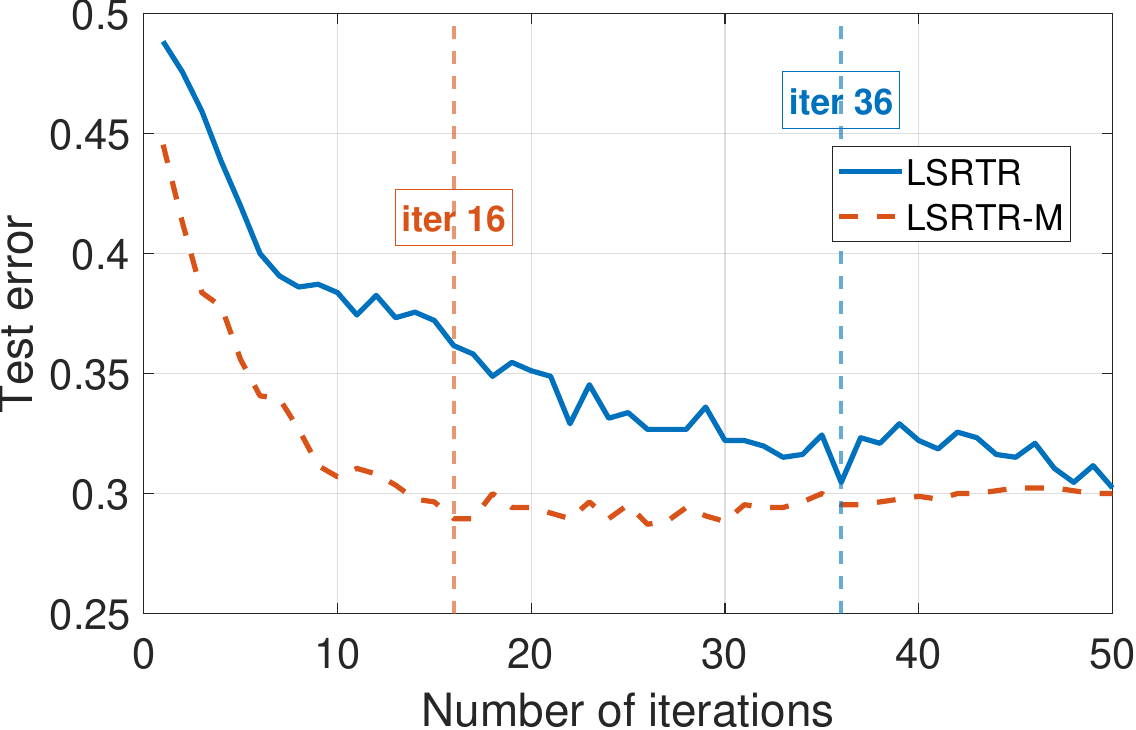}
  \caption{Test error versus number of iterations on the Vessel MNIST 3D dataset under two train-test settings: the original (unbalanced) split (left) and the balanced subsampling setting (right). The dashed vertical lines mark the early-stopping iteration for each method.}
  \label{fig:VesselMNIST_plots}
\end{figure}

We set $\r= (5,5,5)$, $S=3$, and report sensitivity, specificity, F1 score, the Area Under the Curve (AUC), accuracy, and running time. Two evaluation settings are considered:
\begin{itemize}
    \item \emph{Unbalanced split (original).} The training set contains 1335 samples (150 aneurysm / 1185 healthy), and the test set contains 382 samples (43 aneurysm / 339 healthy). Both LSRTR and LSRTR-M are run for 30 iterations. For LSRTR, we set $\alpha=0.7$. For LSRTR-M, we set $\alpha_{m}=0.08$, $\beta=0.3$, and $\lambda=0.05$.
    \item \emph{Balanced split (subsampling).} We construct class-balanced subsets by uniform sampling within each split (150/150 in training and 43/43 in testing). Both algorithms are run for 50 iterations with $\alpha=0.5$, $\alpha_{m}=0.07$, $\beta=0.1$, and $\lambda=0.01$. 
\end{itemize}
For both settings, we first average the error trajectories over 10 independent trials and then apply an early-stopping rule to the resulting mean curve to select a checkpoint for each method.
The dashed vertical lines in Figure~\ref{fig:VesselMNIST_plots} indicate the selected early-stopping iterations. The reported metrics in Tables~\ref{tab:vesselmnist3d_unbalanced} and~\ref{tab:vesselmnist3d_balanced} are evaluated at these selected checkpoints. 

Figure~\ref{fig:VesselMNIST_plots} presents the corresponding test-error trajectories under the unbalanced and balanced settings. In both settings, LSRTR-M consistently achieves lower test error and generally reaches its early stopping point earlier than LSRTR.
Tables~\ref{tab:vesselmnist3d_unbalanced} and~\ref{tab:vesselmnist3d_balanced} summarize the classification performance under the unbalanced and balanced splits, respectively. Across both settings, LSRTR-M consistently achieves improved or competitive classification metrics, including sensitivity, F1 score, AUC, and overall accuracy, while substantially reducing computational time (by approximately 50$\%$). These results demonstrate that the Muon-based update not only accelerates training but also enhances predictive performance on real 3D medical imaging data, further supporting the advantages observed in the synthetic experiments.

\begin{table}[t]
\centering
\footnotesize
\begin{tabular}{lcc}
\hline
Metric & LSRTR & LSRTR-M \\
\hline
Sensitivity & 0.2744 & 0.3395 \\
Specificity & 0.8407 & 0.8493 \\
F1 Score & 0.2168 & 0.2681 \\
AUC & 0.5379 & 0.6232 \\
Accuracy & 0.7770 & 0.7919 \\
Running time (s) & 23.47 & 12.53 \\
\hline
\end{tabular}
\caption{Performance comparison on the Vessel MNIST 3D dataset under the original (unbalanced) split. Metrics are evaluated at the early-stopping iteration of each method.}
\label{tab:vesselmnist3d_unbalanced}
\end{table}

\begin{table}[t]
\centering
\footnotesize
\begin{tabular}{lcc}
\hline
\textbf{Metric} & \textbf{LSRTR} & \textbf{LSRTR-M} \\
\hline
Sensitivity    & 0.7326 & 0.7558 \\
Specificity    & 0.6581 & 0.6651 \\
F1 Score       & 0.7059 & 0.7222 \\
AUC            & 0.7540 & 0.7655 \\
Accuracy       & 0.6953 & 0.7105 \\
Running time (s) & 8.27 & 3.68 \\
\hline
\end{tabular}
\caption{Performance comparison on the Vessel MNIST 3D dataset under the balanced split. Metrics are evaluated at the early-stopping iteration of each method.}
\label{tab:vesselmnist3d_balanced}
\end{table}

\section{Conclusion}
\label{sec:Conclusion}

We introduced LSRTR-M, a Muon-based variant of LSRTR for Low Separation Rank tensor generalized linear models (LSR-TGLMs), with the goal of enabling more efficient estimation from tensor-valued signal and imaging data. The proposed method preserves the original block coordinate descent framework and the core-tensor update of LSRTR, while replacing the projection-based factor-matrix updates with Muon-style orthogonalized momentum steps. This modification eliminates repeated explicit projections and yields more efficient search directions. Extensive experiments on synthetic TGLMs, including linear, logistic, and Poisson regression, demonstrate that LSRTR-M consistently achieves faster convergence in terms of both iteration count and wall-clock time, while achieving lower normalized estimation and prediction errors compared to the baseline LSRTR method. We further evaluated the method on the Vessel MNIST 3D tensor logistic regression task, where LSRTR-M achieves faster training and improved classification performance. 
Overall, the results support LSRTR-M as a computationally efficient approach for structured regression and classification from multidimensional measurements, particularly in settings such as biomedical imaging where preserving tensor structure is important. Future directions include establishing convergence and statistical guarantees for these Muon-type block updates under the LSR structure, designing adaptive step-size and regularization schemes with theoretical support, and extending the framework to larger-scale tensor data and more general structured tensor models.

\section*{Acknowledgment}
This work was supported in part by NSF grants ECCS-2409702.


\bibliographystyle{elsarticle-num}
\bibliography{cas-refs}







\end{document}